\documentclass{article}

\usepackage[preprint]{corl_2026} 

\usepackage{amsmath,amssymb,amsfonts}
\usepackage{graphicx}
\usepackage{booktabs}
\usepackage{multirow}
\usepackage{xcolor}
\usepackage{algorithm}
\usepackage{algorithmic}
\usepackage{subcaption}
\usepackage{wrapfig}
\usepackage{enumitem}

\definecolor{gracolor}{HTML}{F2FBEE}
\definecolor{deltagreen}{HTML}{54B584}
\definecolor{deltared}{HTML}{C84C4C}
\newcommand{\dlt}[1]{\textcolor{deltagreen}{\scriptsize\,#1}}
\newcommand{\dltn}[1]{\textcolor{deltared}{\scriptsize\,#1}}

\newcommand{\ours}{\textbf{GRA}}

\newcommand{\Lsyn}{\mathcal{L}_{2D}^{s}}
\newcommand{\Lreal}{\mathcal{L}_{2D}^{r}}
\newcommand{\Lact}{\mathcal{L}_{\mathrm{act}}}
\newcommand{\Dsyn}{\mathcal{D}_{\mathrm{syn}}}
\newcommand{\Dreal}{\mathcal{D}_{\mathrm{real}}}
\newcommand{\hd}{h_{2\mathrm{D}}}
\newcommand{\phiv}{\phi_v}

\usepackage{enumitem}
\usepackage{xcolor}
\definecolor{citecolor}{HTML}{0071bc}
\definecolor{googleblue}{HTML}{4285F4}
\definecolor{googlered}{HTML}{EA4335}
\usepackage[table]{xcolor}
\definecolor{lightgreen}{RGB}{240, 251, 237}
\definecolor{lightblue}{RGB}{237, 245, 251}
\definecolor{lightred}{RGB}{251, 240, 237}
\definecolor{lightgrey}{RGB}{240,240,240}

\hypersetup{
    colorlinks=true,
    breaklinks=true,
    citecolor=citecolor,
    urlcolor=googleblue,
}

\usepackage{cleveref}

\title{Supervise What Survives: Geometry-Guided VLA Adaptation from Synthetic Robot Videos}

\author{
  \textbf{
  Danze Chen\hspace{1.5em}
  Yanzhe Chen\hspace{1.5em}
  Qiming Huang\hspace{1.5em}
  Zhijun Cao
  }\\[0.1em]
  \textbf{
  Chen Gao\hspace{1.5em}
  Mike Zheng Shou$^\dagger$
  }\\[0.4em]
  Show Lab, National University of Singapore
}

\begin{document}
\begingroup
\renewcommand{\thefootnote}{\fnsymbol{footnote}}
\maketitle
\footnotetext[2]{Corresponding author: mike.zheng.shou@gmail.com}
\endgroup

\vspace{-0.5cm}

\begin{abstract}
Vision-Language-Action (VLA) models require large-scale video-action pairs, yet real teleoperation remains scarce. While generated robot videos offer a scalable alternative, existing methods treat them as real robot data by recovering pseudo-actions from synthesized pixels. We argue that deriving low-level control from generated visuals is a mismatched abstraction.
A video captures only \emph{geometry}: the spatial trajectory representing the \emph{where} of a task. A real demonstration captures \emph{control}: the exact motor commands representing the \emph{how}. Human-to-robot video generation preserves these unequally: the visible geometry survives the generation process, while the underlying control signals are lost. This \textbf{Asymmetric Preservation Principle} dictates a clean rule: this surviving geometry should solely supervise visual perception, leaving control to real demonstrations.
Following this principle, we propose \textbf{GRA} (\textbf{G}eometry-guided \textbf{R}epresentation \textbf{A}lignment), which extracts the geometric content as future 2D end-effector waypoints, computed from the source human video through pose estimation, retargeting, simulation, and calibrated projection, and routes them to the VLA vision backbone via an auxiliary 2D head. The action head is trained on real demonstrations only. During fine-tuning, the waypoint loss persists as a \textbf{spatial representation anchor} that prevents the backbone from losing its geometric grounding.
On real-robot tasks, GRA outperforms pseudo-action baselines under matched data budgets and narrows the gap to policies trained with substantially more real demonstrations, suggesting that correctly routed geometry bridges generated videos to robot policies more reliably than recovered actions.

\end{abstract}



\section{Introduction}
\label{sec:intro}

Vision-Language-Action (VLA) models~\cite{openvla,pi0,pi05,rt2} unify perception and control but face a severe data bottleneck due to the scarcity of paired video-action data~\cite{droid,rtx}: real teleoperation requires dedicated hardware, trained operators, and substantial collection time. Human manipulation videos are abundant, and recent human-to-robot video generation methods~\cite{mitty,phantom,mimicdreamer} make it increasingly feasible to synthesize robot execution videos at scale. However, while these models synthesize visual sequences, they cannot synthesize physical actions. This missing half of the data equation forces us to re-evaluate the exact information these synthetic videos contain, raising a critical question: \emph{what supervision can purely visual generated videos reliably provide, and where should it enter the VLA?}

\begin{figure}[ht]
    \centering
    \includegraphics[width=1\linewidth]{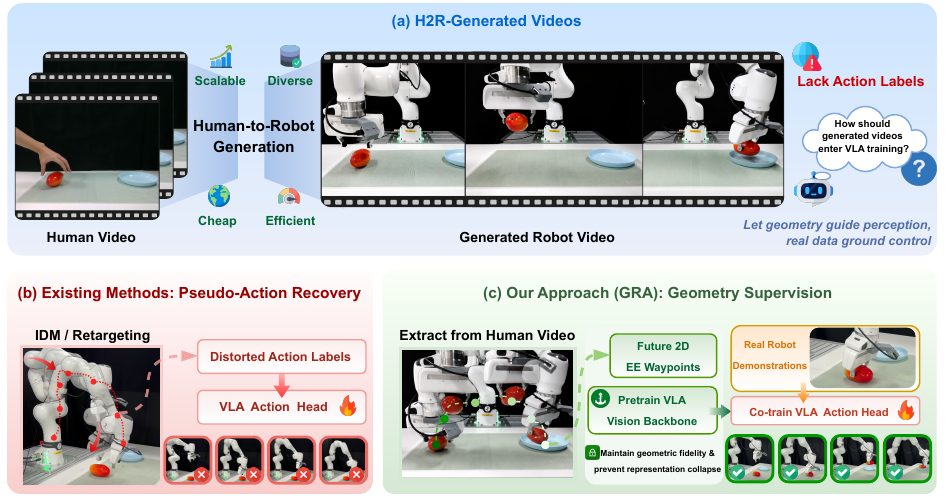} 
    \caption{\textbf{Illustration of Motivation.} \textbf{(a)} Human-to-robot generation produces robot videos at scale, but the generated videos lack action labels. \textbf{(b)} Existing methods recover pseudo-actions from the generated pixels and feed them to the action head, inheriting distorted control signals. \textbf{(c)} \textbf{\ours{}} routes future 2D end-effector waypoints, extracted from the source human video, to the vision backbone, while the action head trains on real robot demonstrations alone.}
    \label{fig:teaser}
    \vspace{-13pt}
\end{figure}

Existing methods answer this by treating generated videos as demonstrations missing their action labels. Prior work recovers pseudo-actions from synthesized frames using inverse dynamics~\cite{dreamgen} or retargeting~\cite{mimicdreamer}, feeding the resulting pairs into the VLA's action head~\cite{gr00tn1_2025}. We argue this creates a fundamental \textbf{video-action mismatch}. Video generation optimizes for visual plausibility, not physical control laws. Forcing the action head to map visually hallucinated pixels to rigid real-world control signals conflates features with physical execution, inevitably degrading policy performance.

To resolve this mismatch, we must analyze the generation process itself. A video captures only \emph{geometry}, namely the spatial structure (end-effector trajectories, object arrangements, and coarse motion directions) representing \emph{where} manipulation unfolds. A real demonstration captures \emph{control}, which consists of the exact motor commands representing \emph{how} to execute it. Video generation is fundamentally a visual synthesis process. It captures the appearance of manipulation, not the physics that drives it. The human-to-robot generation process therefore transfers visible geometric structure reliably, but loses the underlying control signals. We formalize this as the \textbf{Asymmetric Preservation Principle} and derive a simple design rule: \emph{supervision should follow what survives, rather than reconstruct what is erased.} Geometry should guide perception; real demonstrations should ground control.

We propose \textbf{GRA} (\textbf{G}eometry-guided \textbf{R}epresentation \textbf{A}lignment). It routes geometric supervision to the VLA vision backbone via an auxiliary 2D head. Future 2D end-effector waypoints are computed from the source human video through pose estimation, retargeting, physics simulation, and calibrated camera projection, then used to supervise the backbone on generated observations. The action head trains exclusively on real demonstrations. During fine-tuning, the waypoint objective persists as a \textbf{spatial representation anchor}, preventing the backbone from drifting toward action-predictive features at the cost of spatial structure.
On real-robot Franka tasks, GRA outperforms pseudo-action baselines under matched data budgets and narrows the gap to policies trained on real robot data alone with substantially more demonstrations.

Our contributions are summarized as follows:
\vspace{-0.15cm}
\begin{itemize}[leftmargin=*]
    \item \textbf{Asymmetric Preservation.} We formulate the \textbf{Asymmetric Preservation Principle}: generated videos preserve spatial structure more reliably than control dynamics, and should be treated as geometric witnesses rather than control traces.
    \item \textbf{Supervision Routing.} We propose \textbf{GRA}, which routes human-derived geometry to the VLA backbone as future 2D waypoint supervision, independent of synthesized pixels, while learning actions only from real demonstrations.
    \item \textbf{Representational Anchoring.} We demonstrate that action fine-tuning can erode geometry-aware representations; a persistent \textbf{spatial representation anchor} mitigates this drift and is essential to GRA's effectiveness.
\end{itemize}

\section{Related Work}
\label{sec:related}

\paragraph{VLA Models and Adaptation}
VLA models~\cite{openvla,pi0,pi05,rt2,gemini_robo,fast,evolve,dreamvla,qwenrobotmanip,actionmap} couple large-scale vision-language pretraining with action prediction and benefit from cross-embodiment data~\cite{rtx,droid,bridgev2}. Early designs cast actions as discrete tokens within an autoregressive multimodal model~\cite{openvla,rt2}, while more recent variants adopt continuous regression or diffusion-based heads for higher-frequency control~\cite{openvla-oft,pi0,pi05}. Adapting these models to specific hardware still relies on real teleoperation, which is costly to collect at scale. Recent work therefore investigates parameter-efficient fine-tuning~\cite{openvla-oft,lora} and principled data allocation~\cite{aca}. We address an orthogonal axis: how to consume \emph{synthetic} robot videos so that real-data demand is reduced rather than diluted.

\paragraph{Synthetic Data for VLA}
Two sources of synthetic robot data dominate. Simulation provides paired state-action data via domain randomization~\cite{simtoreal_dr} or via demo-conditioned generation~\cite{mimicgen}, but requires task-specific assets. Human-to-robot video generation~\cite{wan22,mitty,phantom,h2r} instead synthesizes robot footage from human demonstrations at scale, with no action labels. The dominant paradigm consumes the generated frames by first recovering pseudo-actions, via inverse dynamics~\cite{dreamgen}, latent action tokens~\cite{moto,lapa}, or geometric retargeting~\cite{mimicdreamer,gr00tn1_2025,gen_robo_pi,videomimic}, and then training the action head on the pseudo-labelled frames. A smaller line instead extracts non-action supervision~\cite{masquerade,hamster,vpp,being,lucibot}: 2D waypoint regression from edited human videos, sparse path tokens, future-frame prediction, and hand-centric video pretraining. GRA belongs to this second line and applies it to fully synthesized robot videos under the Asymmetric Preservation Principle, with a persistent anchor that maintains geometric structure during action fine-tuning.

\paragraph{Representation Preservation in Fine-tuning}
Task-specific fine-tuning risks catastrophic forgetting of pretrained structure~\cite{ewc}. In VLA systems, action-regression gradients can specifically reshape vision features away from spatial understanding, eroding the spatial priors that downstream control depends on. Mitigations include parameter-efficient updates that limit drift~\cite{lora,openvla-oft}, spatial grounding objectives that align visual features with task-relevant geometry~\cite{spatialvla,r3m,voltron,mvp,internvla}, and structured auxiliary supervision that anchors the representation throughout fine-tuning~\cite{code2video,dontblind}. GRA falls into the third category: its spatial anchor uses projected end-effector waypoints as a task-relevant geometric signal, preserving the spatial representations that geometric pretraining builds.

\section{Method}
\label{sec:method}

\begin{figure}[ht]
    \centering
    \includegraphics[width=\columnwidth, trim=0 0 50 0, clip]{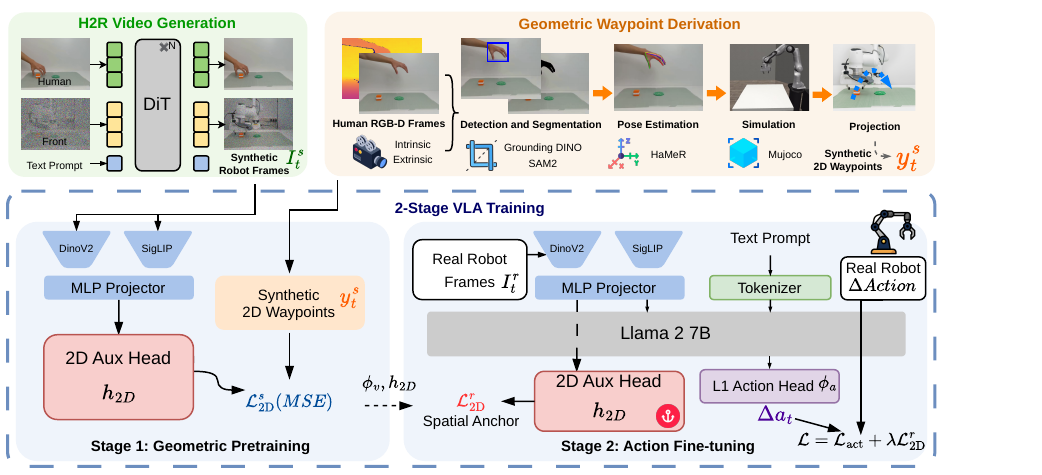}
    \caption{\textbf{Overview of \ours{}.} \textbf{Top:} a human-to-robot video generator produces a robot frame $I_t^s$ from a human demonstration; an independent geometric path extracts 2D end-effector waypoints $y_t^s$ from the source human video. \textbf{Stage 1:} the vision backbone $\phi_v$ and a 2D auxiliary head $h_{2D}$ are trained on $(I_t^s, y_t^s)$ to predict future waypoints under $\Lsyn$. \textbf{Stage 2:} the action head $\phi_a$ is added and the policy is fine-tuned on real robot data with $\mathcal{L}=\Lact+\lambda \Lreal$; $h_{2D}$ continues from Stage~1 as a spatial representation anchor.}
    \label{fig:method}
    \vspace{-13pt}
\end{figure}

GRA realizes the Asymmetric Preservation Principle through a two-stage protocol (Fig.~\ref{fig:method}): a vision backbone is first trained on generated frames with 2D waypoint targets, and an action head is then fine-tuned on real demonstrations while a persistent waypoint loss maintains the spatial structure built in Stage~1. Waypoint targets are derived from the source human video, decoupled from the generated pixels.

\subsection{Task Setup}
\label{sec:notation}

We build on OpenVLA-OFT~\cite{openvla-oft}, which extends OpenVLA~\cite{openvla} with a continuous-action $L_1$-regression head. The policy contains a vision backbone $\phi_v$ that maps an input image to visual tokens in the language-model embedding space, and an action head $\phi_a$ that consumes these tokens after a Llama-2 7B language model conditioned on the task instruction $l$. The policy outputs chunks of $L=24$ delta-action steps $\Delta a_t \in \mathbb{R}^7$, encoding relative end-effector translation, roll-pitch-yaw rotation, and a gripper command, all normalized to $[-1, 1]$. We use $K=8$ as the future-waypoint horizon for spatial supervision; $L$ and $K$ are independent.

The spatial supervision target at frame $t$ is the future $K$-step 2D end-effector trajectory:
\begin{equation}
    y_t = (p_{t+1}, p_{t+2}, \ldots, p_{t+K}), \quad p_\tau \in \mathbb{R}^2,
\end{equation}
where $p_\tau$ is the end-effector position in normalized image coordinates. Training uses two pools: $\Dsyn = \{(I_t^s, y_t^s)\}$ of generated robot frames with waypoint labels (Sec.~\ref{sec:geometric}), and $\Dreal = \{(I_t^r, l, \Delta a_t, y_t^r)\}$ of real teleoperation trajectories with delta actions and projected waypoints. Generated frames are produced by Wan~2.2~\cite{wan22} conditioned on third-person human demonstrations.

\subsection{Geometric Waypoint Derivation}
\label{sec:geometric}

Waypoint targets $y_t^s$ are produced from the \emph{source human video} and a physics simulator, without reading any generated pixel:
\begin{equation}
    \Phi: V^{\text{src}} \longrightarrow (p_1, p_2, \ldots, p_T), \qquad \Phi = \pi_{\text{cam}} \circ \text{FK} \circ \text{IK} \circ \text{HaMeR} \circ \text{Detect}.
\end{equation}
Grounding DINO~\cite{groundingdino} detects hands; SAM2~\cite{sam2} refines bounding boxes into segmentation masks; HaMeR~\cite{hamer} estimates 3D hand keypoints. The 3D hand trajectory is retargeted to the robot frame (wrist$\to$EE position, palm normal$\to$EE orientation, thumb-index distance$\to$gripper width) and replayed in MuJoCo~\cite{mujoco} on a virtual Franka twin to enforce joint limits and collisions. Future $K$-step 3D EE positions are then projected to image coordinates with calibrated camera intrinsics and extrinsics, yielding $y_t^s$.

The supervision target $y_t^s$ and the visual input $I_t^s$ thus originate from separate sources, but they remain temporally aligned: the generated robot video preserves the frame indexing of the source human demonstration, so the waypoint label derived from source frame $t$ is paired with the generated robot frame $I_t^s$ at the same index. For $\Dreal$, waypoints $y_t^r$ are obtained by projecting recorded proprioceptive EE positions through the same camera model.

\subsection{Stage 1: Geometric Pretraining}
\label{sec:stage1}

Stage~1 trains the vision side of the policy on generated frames with a 2D-waypoint regression objective. A three-layer MLP head $h_{2D}$ takes the mean-pooled projected visual tokens from $\phi_v$ and predicts the future $K$-step 2D waypoints $\hat y_t \in \mathbb{R}^{K\times 2}$:
\begin{equation}
    \Lsyn = \mathbb{E}_{(I_t^s,\, y_t^s) \sim \Dsyn} \big[\, \| h_{2D}(\phi_v(I_t^s)) - y_t^s \|_2^2 \,\big].
\end{equation}
In this stage we update $\phi_v$ via LoRA (rank 32) together with the full $h_{2D}$; the language model and the action head are not in the Stage-1 forward path, so generated frames pass only through the visual stack. Training runs for 5K steps.

\subsection{Stage 2: Action Fine-tuning with Spatial Anchor}
\label{sec:stage2}

Stage~2 fine-tunes the policy on real demonstrations only; generated frames do not enter this stage. The action head $\phi_a$ is added now and learns delta actions from real teleoperation, while the auxiliary head $h_{2D}$ from Stage~1 continues to receive a parallel waypoint signal that anchors the spatial structure already built. The total loss combines $L_1$ delta-action regression with the waypoint loss from Stage~1, now evaluated on real frames:
\begin{equation}
    \mathcal{L} = \Lact + \lambda \Lreal,
\end{equation}
where
\begin{align}
    \Lact &= \mathbb{E}_{(I_t^r,\, l,\, \Delta a_t) \sim \Dreal} \big[\, \| \phi_a(\phi_v(I_t^r), l) - \Delta a_t \|_1 \,\big], \\
    \Lreal &= \mathbb{E}_{(I_t^r,\, y_t^r) \sim \Dreal} \big[\, \| h_{2D}(\phi_v(I_t^r)) - y_t^r \|_2^2 \,\big].
\end{align}
The waypoint targets $y_t^r$ are obtained from the recorded EE proprioception of the same trajectory, so $\Lreal$ adds no extra labelling cost on top of teleoperation. We set $\lambda = 0.5$ and apply LoRA (rank 32) to $\phi_v$ and the language model; $h_{2D}$ is initialised from its Stage-1 weights and $\phi_a$ from random weights, both fully updated. Training runs for 10K steps.

The auxiliary loss $\Lreal$ acts as a \textbf{spatial representation anchor}: it continually re-imposes a waypoint-prediction objective on $\phi_v$ during action fine-tuning, counteracting the drift that action-regression gradients would otherwise induce on the spatial features built in Stage~1. We test this mechanism directly in Sec.~\ref{sec:ablation} by removing $\Lreal$ while keeping the Stage-1 initialisation.

\section{Experiments}
\label{sec:experiments}

We address three questions in real-robot experiments: (i) whether GRA outperforms pseudo-action baselines under a matched real-data budget; (ii) what mechanism drives the gain; and (iii) which GRA components contribute.

\subsection{Setup}
\label{sec:setup}

\begin{figure}[ht]
    \centering
    \includegraphics[width=\linewidth]{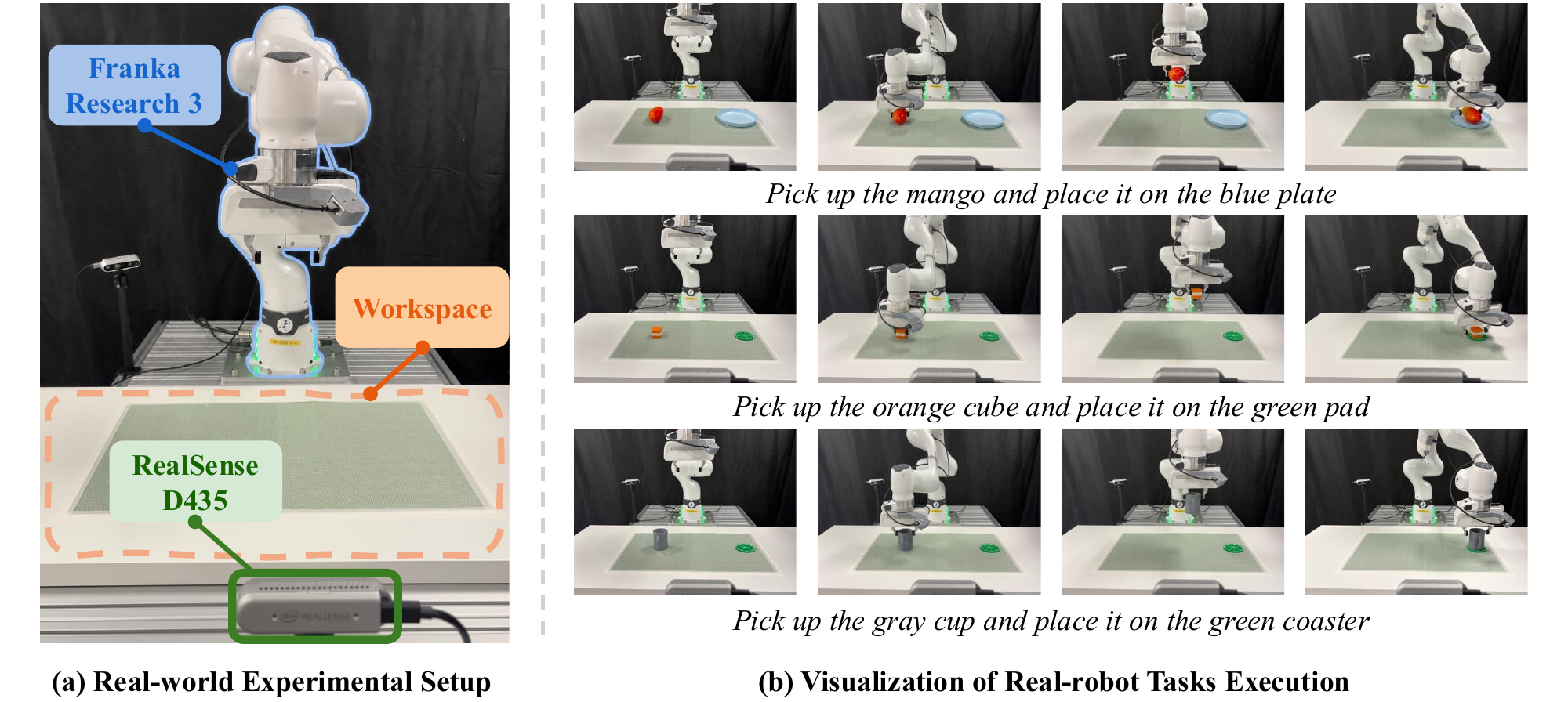}
    \caption{Illustration of (a) the real-robot experimental setup and (b) rollout examples of the three pick-and-place tasks.}
    \label{fig:setup}
    \vspace{-7pt}
\end{figure}

\textbf{Hardware \& Tasks.}
All experiments are performed on a 7-DoF Franka Research 3 with a fixed third-person RealSense D435 camera. We evaluate three tabletop pick-and-place tasks (\emph{cube$\to$pad}, \emph{cup$\to$coaster}, and \emph{mango$\to$plate}), each evaluated over 30 trials. We report per-task success rate and the 3-task mean. The setup and the three tasks are illustrated in Fig.~\ref{fig:setup}.

\textbf{Data.}
For each task we use $N_{\text{real}}{=}25$ real teleoperated trajectories and $N_{\text{syn}}{=}75$ generated videos. Generated videos are produced by Wan~2.2~\cite{wan22} conditioned on third-person human videos, at 480$\times$640 resolution, 81~frames.

\textbf{Baselines.}
All baselines use the OpenVLA-OFT architecture~\cite{openvla-oft} with identical hyperparameters (batch size 16, learning rate $5{\times}10^{-4}$, LoRA rank 32). \textbf{Real-only}: 10K steps on $\Dreal$ with $N_{\text{real}}{=}25$. \textbf{Real-only (full)}: same recipe with $N_{\text{real}}{=}100$, serving as an upper-reference. \textbf{DreamGen-style}~\cite{dreamgen}: IDM-recovered pseudo delta-actions on generated frames jointly trained with real data, 10K steps. \textbf{MimicDreamer-style}~\cite{mimicdreamer}: delta-action labels obtained by geometric retargeting on generated frames, 10K steps of joint behaviour cloning. \textbf{GRA (ours)}: 5K Stage~1 pretraining + 10K Stage~2 cotraining; $K{=}8$, $\lambda{=}0.5$.

\subsection{Main Results}
\label{sec:main_results}

\begin{table*}[ht]
\centering
\caption{\textbf{Main results.} Closed-loop success rate on three real-robot pick-and-place tasks. We report \emph{successful trials} and the corresponding \textit{success rate} (\%). The Mean column shows the 3-task average; the small coloured number is the absolute change vs.\ the matched-budget Real-only baseline. Best per task within the matched-budget setting in \textbf{bold}.}
\label{tab:main}
\vspace{0.3em}
\setlength{\tabcolsep}{6.5pt}
\renewcommand{\arraystretch}{1.25}
\begin{tabular}{l cc cc cc cc}
\toprule
\multirow{2}{*}{Method} &
\multicolumn{2}{c}{\textbf{Cube $\to$ Pad}} &
\multicolumn{2}{c}{\textbf{Cup $\to$ Coaster}} &
\multicolumn{2}{c}{\textbf{Mango $\to$ Plate}} &
\multicolumn{2}{c}{\textbf{Overall}} \\
\cmidrule(lr){2-3}\cmidrule(lr){4-5}\cmidrule(lr){6-7}\cmidrule(lr){8-9}
 & Succ. & Rate & Succ. & Rate & Succ. & Rate & Rate & + / - \\
\midrule
\rowcolor{gray!12}\multicolumn{9}{l}{\emph{Upper reference:} $N_{\text{real}}{=}100$} \\
Real-only (full)     & 22 & 73.3 & 19 & 63.3 & 27 & 90.0 & 75.6 & \dlt{+14.5} \\
\midrule
\rowcolor{gray!12}\multicolumn{9}{l}{\emph{Matched-budget setting:} $N_{\text{real}}{=}25$, $N_{\text{syn}}{=}75$} \\
Real-only            & 18 & 60.0 & 14 & 46.7 & 23 & 76.7 & 61.1 & \\
DreamGen-style       & 14 & 46.7 & 11 & 36.7 & 19 & 63.3 & 48.9 & \dltn{$-$12.2} \\
MimicDreamer-style   & 16 & 53.3 & 12 & 40.0 & 21 & 70.0 & 54.4 & \dltn{$-$6.7}  \\
\rowcolor{gracolor}
\textbf{GRA (ours)}  & \textbf{20} & \textbf{66.7} & \textbf{17} & \textbf{56.7} & \textbf{25} & \textbf{83.3} & \textbf{68.9} & \dlt{+7.8} \\
\bottomrule
\end{tabular}
\end{table*}

\textbf{(i) Geometric routing outperforms pseudo-action labelling.} GRA achieves the highest matched-budget mean ($68.9\%$, $+7.8$\,pt over Real-only) and is best on all three tasks, while using generated videos only as geometric supervision for the visual backbone.

\textbf{(ii) Pseudo-action supervision hurts in our setting.} Both pseudo-action baselines fall below Real-only ($-12.2$\,pt for DreamGen-style, $-6.7$\,pt for MimicDreamer-style), consistently across all three tasks. This is consistent with the precision sensitivity of delta actions, where small pseudo-label biases can compound during closed-loop execution.

\textbf{Closing the gap to the full-data reference.} GRA narrows the gap to the 100-demo Real-only reference from $14.5$\,pt to $6.7$\,pt while using $4\times$ fewer real demonstrations.

\subsection{Diagnosing the Information Asymmetry}
\label{sec:diagnostics}

We examine \emph{why} through three diagnostics that follow the data path: what generated frames preserve, what the trained backbone retains, and how the action head behaves.

\textbf{Preservation gap (data).}
We first measure how spatial vs.\ control information transfers from generated to real frames. We instantiate two frozen encoders: (i) an off-the-shelf SigLIP, as a no-fine-tuning reference; (ii) the GRA Stage-1 backbone, after our 5K-step waypoint pretraining. On top of each, we train two lightweight MLP probes (2-layer, 256 hidden units): a waypoint predictor (single-frame features $\to$ $2K$-dim output) and a delta-action predictor (two consecutive frames concatenated $\to$ 7-dim output). Each probe is trained with real frames as input ($\mathrm{R}\!\to\!\mathrm{R}$, in-domain) and with generated frames as input ($\mathrm{S}\!\to\!\mathrm{R}$, transfer); both variants are evaluated on the same real-validation split. Since waypoint and $\Delta$action targets live on different scales, we report normalized $L_1$ error in $\sigma$-units of validation labels (smaller is better).

\begin{figure}[ht]
    \centering
    \includegraphics[width=0.9\linewidth]{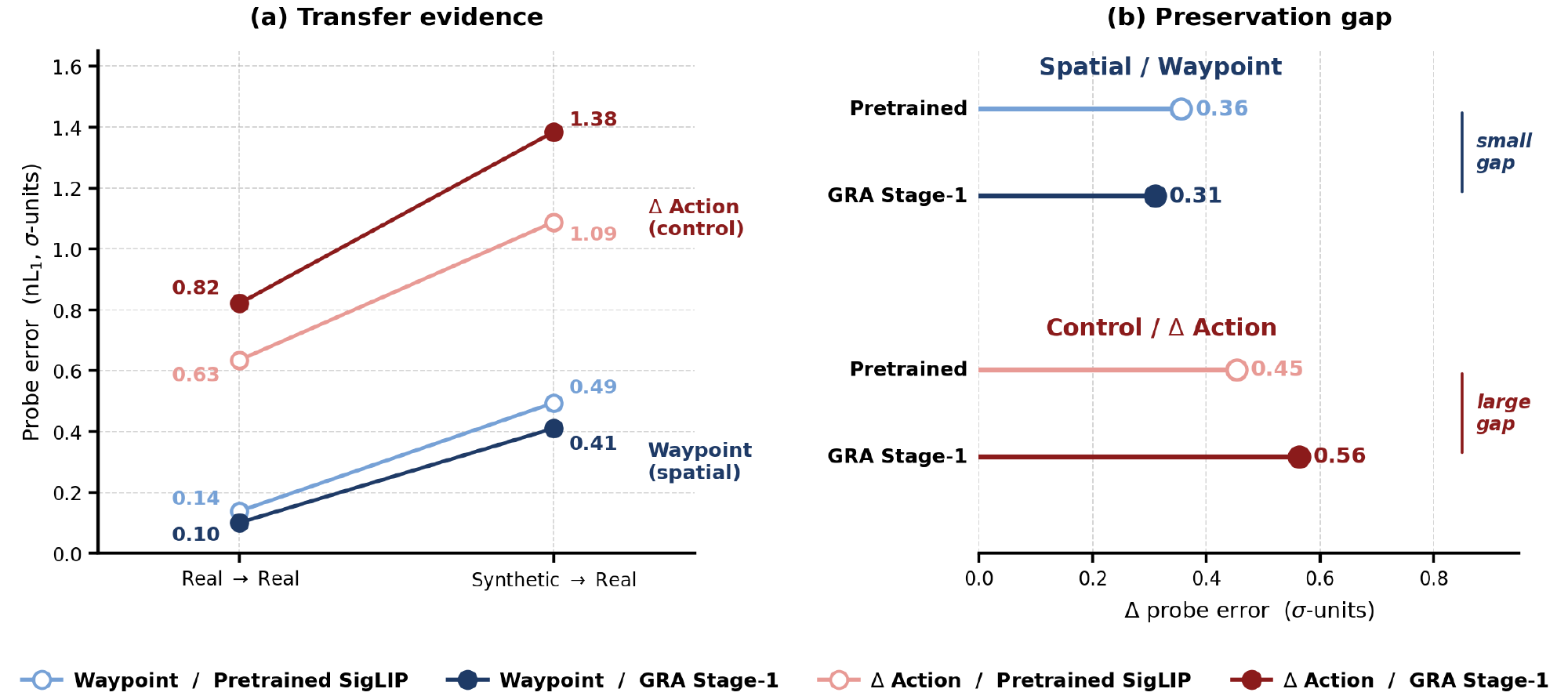}
    \caption{Information preservation gap measured with frozen vision features (mean over three tasks). \textbf{(a)} Probe error under in-domain ($\mathrm{R}\!\to\!\mathrm{R}$) and generated-to-real ($\mathrm{S}\!\to\!\mathrm{R}$) evaluation for waypoint and $\Delta$action targets. \textbf{(b)} Preservation gap, $\mathrm{nL_1}(\mathrm{S}\!\to\!\mathrm{R}) - \mathrm{nL_1}(\mathrm{R}\!\to\!\mathrm{R})$; lower is better.}
    \label{fig:preservation}
    \vspace{-5pt}
\end{figure}

Three patterns emerge from Fig.~\ref{fig:preservation}. \textbf{(i) Native asymmetry in vision features.} On the off-the-shelf SigLIP, the in-domain probe error is $0.14\sigma$ for waypoints versus $0.63\sigma$ for $\Delta$actions; spatial structure is approximately $4\times$ more readily recoverable, even before any fine-tuning. \textbf{(ii) Transfer asymmetry.} A syn-trained waypoint probe attains $0.41\sigma$ on real frames (only $0.31\sigma$ above its in-domain bound on the Stage-1 backbone), whereas a syn-trained $\Delta$action probe degrades to $1.38\sigma$ on real frames, more than $3\times$ the waypoint error and exceeding the in-domain $\Delta$action error itself; this indicates substantially weaker cross-domain transfer for control than for spatial waypoints. \textbf{(iii) Routing asymmetry.} Stage-1 waypoint pretraining further \emph{reduces} in-domain waypoint error (to $0.10\sigma$) yet \emph{increases} $\Delta$action error (to $0.82\sigma$), suggesting that the supervision route reshapes which information remains accessible in the backbone.

\textbf{Spatial probing (backbone).}
Having shown the asymmetry exists in the data, we next test whether the trained vision backbone retains accessible spatial structure. After training, we freeze each method's $\phi_v$ and fit an MLP probe (mean-pooled features $\to$ 2D current end-effector pixel location) on held-out real frames. Since no method directly supervises this projection during training, the probe measures intrinsic spatial perception. We report mean $\pm$ standard deviation of $L_1$ error over five random splits in the left columns of Tab.~\ref{tab:diagnostics}. As a no-fine-tuning reference, the off-the-shelf SigLIP encoder achieves $0.039 \pm 0.004$ ($25.7\%$ within $10$\,px), confirming that all trained backbones add task-relevant spatial features beyond generic pretraining.

\begin{table}[ht]
\centering
\small
\caption{\textbf{Diagnostic measurements} on the trained vision backbone and the action head. \emph{Left:} spatial-readout $L_1$ pixel error (mean$\pm$std over five random splits; lower $L_1$ / higher \%${<}$10\,px is better). \emph{Right:} per-step action error (lower is better). Best per column in \textbf{bold}.}
\label{tab:diagnostics}
\setlength{\tabcolsep}{6pt}
\begin{tabular}{l cc cc}
\toprule
 & \multicolumn{2}{c}{\textbf{Spatial readout (frozen probe)}}
 & \multicolumn{2}{c}{\textbf{Per-step action error}} \\
\cmidrule(lr){2-3}\cmidrule(lr){4-5}
Method & $L_1$ $\downarrow$ & \%${<}$10\,px $\uparrow$ & Pos.\ (mm) $\downarrow$ & Total $L_1$ $\downarrow$ \\
\midrule
Real-only           & 0.022 $\pm$ 0.003 & 51.3 & 11.09 & 0.090 \\
DreamGen-style      & 0.029 $\pm$ 0.018 & 47.7 & 10.04 & 0.080 \\
MimicDreamer-style  & 0.020 $\pm$ 0.003 & \textbf{51.5} & 13.36 & 0.071 \\
\rowcolor{gracolor}
\textbf{GRA (ours)} & \textbf{0.020 $\pm$ 0.002} & 50.4 & \textbf{9.26}  & \textbf{0.060} \\
\bottomrule
\end{tabular}
\end{table}

Two findings stand out from the spatial-readout columns. \textbf{(i) GRA matches the strongest baseline.} GRA's $L_1$ ($0.020$) ties MimicDreamer-style on the mean and is the most consistent across seeds (std $0.002$); the two methods stay within $1.1$\,pt on \%${<}$10\,px. \textbf{(ii) The backbone need not dominate to support strong control.} MimicDreamer-style attains comparable probe accuracy yet its closed-loop success is $14.5$\,pt below GRA (Tab.~\ref{tab:main}). We read this as \emph{designed equivalence rather than dominance}: because GRA externalises spatial supervision into the auxiliary head $h_{2D}$, the backbone reaches probing accuracy on par with the strongest pseudo-action baseline without needing to dominate it.

\textbf{Action head contamination.}
Finally, we test whether pseudo-action supervision actively degrades the action head via a teacher-forcing protocol: for held-out real trajectories, we feed ground-truth observations to each policy and record predicted delta actions at every step, isolating per-step action quality from compounding errors. Per-step results are reported in the right columns of Tab.~\ref{tab:diagnostics}.

Two observations follow from the per-step columns. \textbf{(i) Position drift is reduced by $17\%$} relative to Real-only ($9.26$ vs.\ $11.09$\,mm) and $30\%$ relative to MimicDreamer-style ($13.36$\,mm), showing that geometric pretraining sharpens spatial perception in a way that translates to per-step action accuracy. \textbf{(ii) Total $L_1$ error drops from $0.090$ on Real-only to $0.060$ on GRA}, suggesting that the action head benefits from spatial pretraining while avoiding synthetic pseudo-action noise.

\subsection{Ablation Study}
\label{sec:ablation}

\begin{wraptable}{r}{0.58\textwidth}
    \vspace{-0.4cm}
    \centering
    \caption{\textbf{Component ablations} on \emph{cup$\to$coaster}.}
    \label{tab:ablation}
    \begin{tabular}{lc}
    \toprule
    Variant & SR (\%) \\
    \midrule
    \rowcolor{gracolor}
    \textbf{GRA full}                   & \textbf{56.7} \\
    w/o Stage~1 (anchor only)           & 43.3 \\
    w/o Anchor (drop $\Lreal$)          & 46.7 \\
    w/o Geometric Path (delta-action target) & 36.7 \\
    \bottomrule
    \end{tabular}
    \vspace{-0.05cm}
\end{wraptable}

All baselines achieve their lowest success rates on \emph{cup$\to$coaster} (Tab.~\ref{tab:main}), so we ablate GRA on this task. Tab.~\ref{tab:ablation} shows that all three components contribute. \textbf{(i) Stage~1 pretraining is the foundation.} Removing it (\emph{w/o Stage~1}: initialize from base OpenVLA-7B with $\lambda{=}0.5$ during cotraining) drops success to $43.3\%$, below Real-only ($46.7\%$). \textbf{(ii) The anchor helps preserve Stage-1 structure.} Removing $\Lreal$ (\emph{w/o Anchor}: $\lambda{=}0$ during Stage~2) reduces success to $46.7\%$, suggesting that the persistent waypoint loss helps retain spatial features during action fine-tuning. \textbf{(iii) The supervision content matters.} Replacing waypoint targets with retargeted EE delta-action targets, while keeping the same auxiliary-routing design (\emph{w/o Geometric Path}), performs worst ($36.7\%$, below Real-only), indicating that routing synthetic supervision through the backbone is insufficient if the supervision content is itself control-side noisy.

\section{Limitations and Conclusion}
\label{sec:conclusion}

\textbf{Limitations.}
GRA addresses one part of the synthetic-to-real stack: how generated videos should be routed into a VLA when only part of their signal is trustworthy. Its limits therefore lie both upstream and downstream. Upstream, the spatial-control asymmetry is observed empirically; characterising it across generators and extending ``what survives'' beyond geometry remain open problems. Downstream, closed-loop performance remains tied to the scale of real action data; lifting this ceiling will require future generators or action decoders that yield trustworthy control signals.

\textbf{Conclusion.}
We studied how generated robot videos should enter VLA training. The Asymmetric Preservation Principle identifies a structural gap between what video generation preserves reliably (spatial geometry) and what it fails to preserve reliably (control dynamics). GRA routes synthetic supervision accordingly: geometric targets supervise the vision backbone, while action learning remains grounded in real demonstrations. Under a matched real-data budget, this routing outperforms pseudo-action baselines on three real-robot tasks and narrows the gap to a Real-only policy trained with $4\times$ more demonstrations. More broadly, our results suggest that synthetic-to-real pipelines should supervise what survives generation, rather than reconstruct what generation erases.

\section{Acknowledgement}
\label{sec:Acknowledgement}

We thank Kevin Yuchen Ma and Guian Fang for their support in generating the reference data. We are also grateful to Qi Lv and Xiaokang Liu for their valuable advice and insightful discussions.

\clearpage
\bibliography{references}  

\clearpage
\appendix

\noindent
\section{Implementation Details}
\label{sec:sup-impl}

Table~\ref{tab:sup-hparams} consolidates the settings for the two stages of \ours{} and the three baselines (Real-only, DreamGen-style, MimicDreamer-style). All five runs use the same OpenVLA-OFT backbone~\cite{openvla-oft} and the same optimiser, batch size, learning rate, and LoRA configuration; they differ only in the number of optimisation steps, the parameters that receive gradients, and the supervision route.

\begin{table}[h]
\centering
\small
\caption{Training hyperparameters.}
\label{tab:sup-hparams}
\begin{tabular}{ll}
\toprule
Item & Setting \\
\midrule
Optimiser            & AdamW, $\beta=(0.9,0.999)$, wd $0.01$, bf16 \\
Batch size           & 16 \\
Learning rate        & $5\times10^{-4}$, cosine schedule \\
LoRA rank / alpha    & 32 / 64 \\
Hardware             & 1 $\times$ H200 \\
\midrule
\multicolumn{2}{l}{\textit{Stage~1} (\ours{} pretrain on $\Dsyn$)} \\
\quad Steps          & 5\,K (200 warmup) \\
\quad LoRA on        & $\phiv$ \\
\quad Waypoint horizon $K$ & 8 \\
\midrule
\multicolumn{2}{l}{\textit{Stage~2} (\ours{} cotrain) and baselines} \\
\quad Steps          & 10\,K (500 warmup) \\
\quad LoRA on        & $\phiv$ + LM \\
\quad Action chunk $L$ & 24 \\
\quad Anchor weight $\lambda$ ({\ours} only) & 0.5 \\
\bottomrule
\end{tabular}
\end{table}

The auxiliary spatial head $\hd$ is a 3-layer MLP with hidden dimensions $1024\!\to\!512\!\to\!256$, GELU activations, dropout $0.1$, and a final linear projection to $2K=16$ outputs; Stage~2 cotraining loads its Stage~1 weights as the initialisation and continues to update all of its parameters.

The synthetic robot videos at Stage~1 are produced by a Wan-2.2 model~\cite{wan22} fine-tuned with low-rank adapters under the recipe of Mitty~\cite{mitty}: starting from the released Mitty adapter, the adapter is further fine-tuned on the source human episodes used in this work and then run in inference mode to generate the videos. Fine-tuning and inference both run in bf16 mixed precision under the DiffSynth backend, with the task instruction (\textit{``pick up X and place on Y''}) as the text prompt and conditioning frames sampled from the source human demonstration. A single shared adapter is reused across all three tasks, with only the conditioning frames changing between them; resolution, frame count, and frame rate are reported in the main-paper data description.

\section{Data Pipeline and IDM Implementation}
\label{sec:sup-data}

The five-stage source-to-waypoint pipeline is described in main-paper Sec.~3.2; this section lists the implementation choices that did not fit into the main text. \textbf{Hand detection} uses Grounding~DINO~\cite{groundingdino} with the text prompt \textit{``hand''} at confidence threshold $0.4$, followed by SAM~2~\cite{sam2}, which refines the predicted bounding box into a segmentation mask used as a sanity filter against false positives from background humans. \textbf{Hand reconstruction} runs HaMeR~\cite{hamer} with its ViT-B backbone and the official MANO output head, producing 21 3D joints in the camera frame. \textbf{Retargeting and replay} place the retargeted EE trajectory in MuJoCo with the official Franka MJCF model, which provides the kinematic constraints used during inverse kinematics.

The DreamGen-style baseline additionally requires an inverse-dynamics model that produces per-step pseudo-actions on generated frames, with one such IDM trained per task. The IDM is a frozen SigLIP-base encoder followed by a 4-layer MLP that maps two consecutive 224$\times$224 frames to a 7-dimensional output (position delta, orientation delta, binary gripper). Training uses the real teleoperation episodes for that task, an 80/20 episode-level split, AdamW with learning rate $10^{-3}$ and batch size $64$, and runs for 30 epochs. The trained IDM is then applied frame-by-frame to each generated video, and the resulting per-step pseudo-actions are paired with the corresponding generated frames as the synthetic-side training data for the DreamGen-style baseline.

\section{Evaluation Protocol}
\label{sec:sup-eval}

The hardware setup and the three tasks are described in main-paper Sec.~4.1. At inference time the action chunk size is $L=24$ and the policy is rolled out in a receding-horizon fashion: at each control step the chunk is predicted and only the first action is executed before re-querying. Commands are issued to the robot at $10$\,Hz, and each trial times out at $30$\,s. A trial is counted as a success when the object is placed on or in the target region with the gripper released, judged by the experimenter at trial end; borderline cases—the object touches the target but slips off, or the gripper opens before contact—are recorded as failures. The same experimenter scores all four methods on the same trials.

The three diagnostic probes in main-paper Sec.~4.3 share a common offline evaluation set: held-out real teleoperation trajectories not used in any training. The probes for the \emph{information preservation gap} are 2-layer MLPs with hidden width $256$, GELU activations, and dropout $0.1$, each trained for $5$\,K steps with Adam at learning rate $10^{-3}$; the five random splits used to compute the mean$\pm$std are episode-level, so frames from the same episode never appear in both training and validation, and errors are normalised to $\sigma$-units using the per-target standard deviation on the validation split, which makes waypoint and $\Delta a$ errors comparable on a single axis. The \emph{spatial readout probe} shares this architecture and training schedule and is fitted on top of the frozen mean-pooled visual tokens of each method's vision backbone, with the current 2D EE pixel location as the target. For the \emph{teacher-forcing per-step error}, the position component is reported in millimetres after un-normalising the predicted $\Delta a_t$ back to the original action range; the total $\ell_1$ is the un-weighted $\ell_1$ over the full 7-DoF action.

\section{Stage~1 Spatial Probe: Sanity Check}
\label{sec:sup-stage1}

This section verifies that Stage~1 pretraining converges and that the spatial probe~$\hd$ generalises to held-out episodes from the Stage~1 corpus. For each task, six episodes are held out and evaluated against the geometric ground truth on the predicted $K{=}8$ future EE waypoints. The aggregate errors are reported in Table~\ref{tab:sup-stage1-mae}: all three tasks land within $\sim$3--5\,\% of image width, well below the spatial scale of the manipulation region.

\begin{table}[h]
\centering
\small
\caption{Stage~1 sanity check: held-out waypoint MAE.}
\label{tab:sup-stage1-mae}
\begin{tabular}{lcc}
\toprule
Task  & MAE (px) & Rel.\ (\%) \\
\midrule
Cube  & 23.2 & 4.8 \\
Cup   & 18.5 & 3.8 \\
Mango & 19.7 & 4.1 \\
\bottomrule
\end{tabular}

\end{table}

\begin{figure}[h]
\centering
\includegraphics[width=1\linewidth]{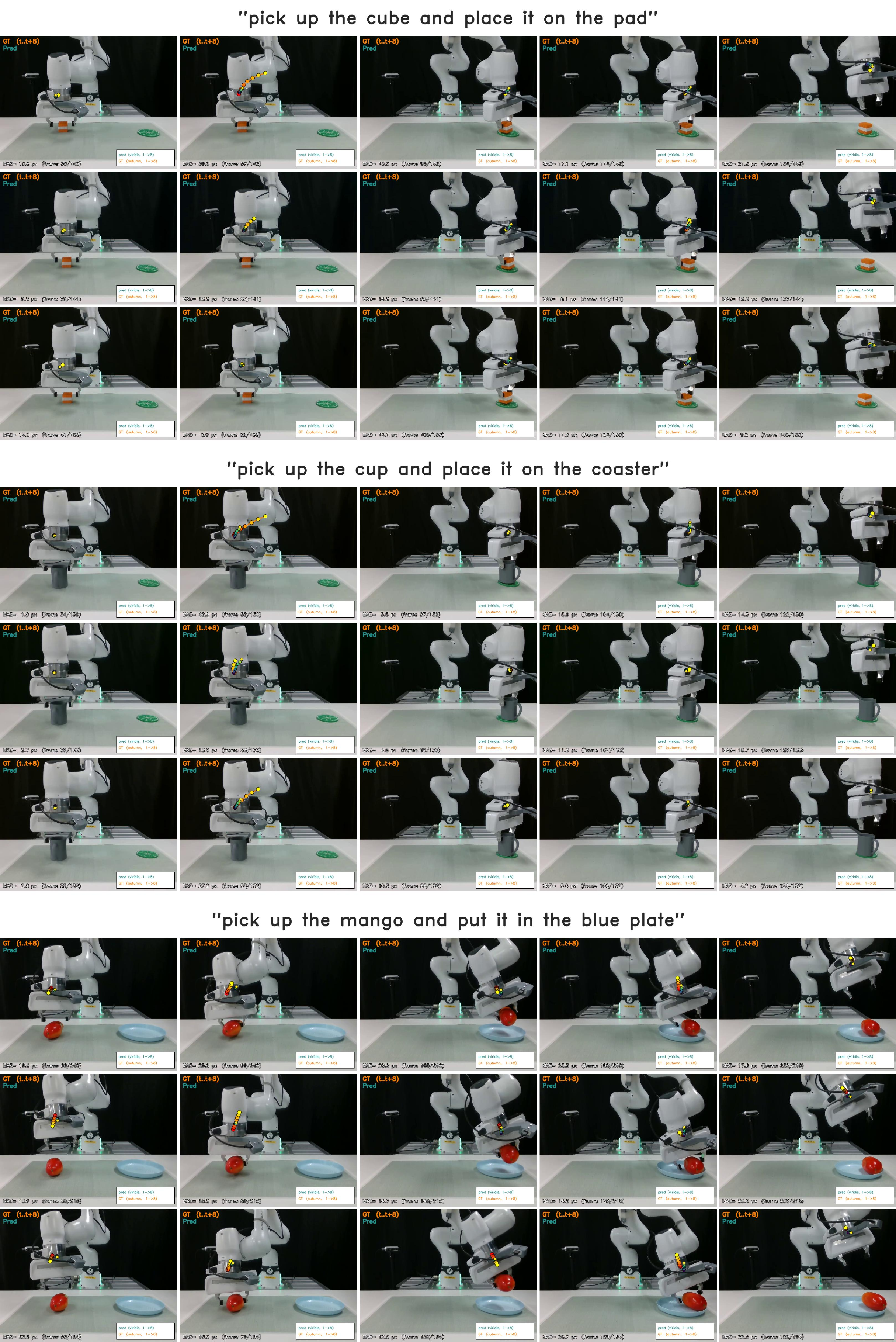}
\caption{\textbf{Stage~1 spatial probe predictions on held-out episodes.} Each task block shows three held-out episodes (rows) at five equi-spaced time steps (columns); the language instruction for the task is given above the block. At every panel the $K{=}8$ future EE waypoints are shown as eight dots connected by a polyline; red is the geometric ground truth from the source-to-waypoint pipeline, blue is the prediction of $\hd$.}
\label{fig:sup-b2}
\end{figure}

To complement the aggregate error, three held-out episodes per task are visualised at five equi-spaced time steps drawn from the contact and post-contact phases of each rollout (Fig.~\ref{fig:sup-b2}). At every panel the $K{=}8$ future EE waypoints are rendered as eight dots connected by a polyline; red marks the geometric ground truth from the source-to-waypoint pipeline, and blue marks the prediction of $\hd$. The blue waypoints stay tightly aligned with red across the approach, contact, and lift phases, including frames in which the gripper partially occludes the wrist.

\end{document}